\documentclass[letterpaper, 10 pt, journal, twoside]{IEEEtran}

\IEEEoverridecommandlockouts                              

\usepackage[utf8]{inputenc} 
\usepackage[T1]{fontenc}    

\usepackage[draft]{hyperref}  

\usepackage{url}            
\usepackage{booktabs}       
\usepackage{amsfonts}       
\usepackage{nicefrac}       
\usepackage{microtype}      
\usepackage{xcolor}         
\usepackage{wrapfig}

\usepackage{float}  

\usepackage{csquotes} 
\usepackage{times}
\usepackage{multicol}
\usepackage{amsmath}
\usepackage{amssymb}
\usepackage{graphicx}
\usepackage[font=small]{caption}
\usepackage{subcaption}
\usepackage{multirow} 
\usepackage{booktabs} 
\usepackage{todonotes}
\usepackage{dblfloatfix}

\newif\ifdiff
\difffalse   

\usepackage{xcolor}
\usepackage[normalem]{ulem} 

\ifdiff
  \newcommand{\add}[1]{\textcolor{blue}{#1}}
  \newcommand{\del}[1]{\textcolor{red}{\sout{#1}}}
\else
  \newcommand{\add}[1]{#1}
  \newcommand{\del}[1]{}
\fi
\newcommand{\diff}[2]{\del{#1}\add{#2}}

\begin{document}
\title{Self-Supervised Multisensory Pretraining for\\ Contact-Rich Robot Reinforcement Learning}

\author{Rickmer Krohn$^{1, 2, 3}$, Vignesh Prasad$^{1, 2, 3}$, Gabriele Tiboni$^{1}$ and Georgia Chalvatzaki$^{1, 2, 3}$
\thanks{ 
       Received 24 October 2025; accepted 20 March 2026. Date of publication 6 April 2026; This article was recommended for publication by Associate Editor P. Falco and Editor J. Kober upon evaluation of the reviewers’ comments. \\ Corresponding author: {\tt\footnotesize rickmer.krohn@tu-darmstadt.de}.}%
\thanks{
        $^{1}$~Interactive Robot Perception \& Learning (PEARL) Lab, TU Darmstadt, Germany,
        $^{2}$ Hessian.AI;  $^{3}$ Robotics Institute Germany (RIG); This research is funded by the German Research Foundation (DFG) Emmy Noether Programme (CH 2676/1-1), the EU’s Horizon Europe project \enquote{ARISE} (Grant no.: 101135959), the German Federal Ministry of Education and Research (BMBF) project “RiG” (Grant no.: 16ME1001) and the European Research Council (ERC) project “SIREN” (Grant No.: 101163933).}
\thanks{ 
         Digital Object Identifier 10.1109/LRA.2026.3681156}%
\thanks{ 
         \copyright 2022 IEEE. Personal use of this material is permitted. Permission
    from IEEE must be obtained for all other uses, in any current or future media,
    including reprinting/republishing this material for advertising or promotional
    purposes, creating new collective works, for resale or redistribution to servers
    or lists, or reuse of any copyrighted component of this work in other works.%
    }   
}

\markboth{IEEE ROBOTICS AND AUTOMATION LETTERS. PREPRINT VERSION. ACCEPTED MARCH, 2026}
{Krohn \MakeLowercase{\textit{et al.}}: Self-Supervised Multisensory Pretraining for Contact-Rich Robot Reinforcement Learning} 





\maketitle

\begin{abstract}
Effective contact-rich manipulation requires robots to synergistically leverage vision, force, and proprioception. However, Reinforcement Learning agents struggle to learn in such multisensory settings, especially amidst sensory noise and dynamic changes. 
We propose MultiSensory Dynamic Pretraining (MSDP), a novel framework for learning expressive multisensory representations tailored for task-oriented policy learning. MSDP is based on masked autoencoding and trains a transformer-based encoder by reconstructing multisensory observations from only a subset of sensor embeddings, leading to cross-modal prediction and sensor fusion. For downstream policy learning, we introduce a novel asymmetric architecture, where a cross-attention mechanism allows the critic to extract dynamic, task-specific features from the frozen embeddings, while the actor receives a stable pooled representation to guide its actions. Our method demonstrates accelerated learning and robust performance under diverse perturbations, including sensor noise, and changes in object dynamics. Evaluations in multiple challenging, contact-rich robot manipulation tasks in simulation and the real world showcase the effectiveness of MSDP. Our approach exhibits strong robustness to perturbations and achieves high success rates on the real robot with as few as 6,000 online interactions, offering a simple yet powerful solution for complex multisensory robotic control. Website: \href{https://msdp-pearl.github.io/}{https://msdp-pearl.github.io/}
    
\end{abstract}

\vspace{-0.5em}
\section{Introduction}
\label{sec-intro}
\IEEEPARstart{R}{einforcement} Learning (RL) has shown impressive successes in learning complex tasks ranging from Atari \cite{mnih_playing_2013}, locomotion \cite{zhuang_robot_2023}, vision-based manipulation \cite{zhang_towards_2015} to multisensory peg insertion \cite{lee_making_2018,sferrazza_power_2023}.
However, incorporating multiple sensor modalities, especially in complex contact-rich robotic manipulation tasks, remains a challenge for RL, due to the heterogeneous dynamics of different sensor modalities. Additionally, the importance of each input modality changes during the execution of a manipulation task, e.g., coarse scene understanding from visual input to fine-grained force feedback when in contact. Thus, robotic agents need to learn how to dynamically focus on the most relevant sensory information while adapting to perturbations and dynamic changes in the environment. 
This \diff{capability, referred to as}{field of} \emph{sensor fusion}\diff{,}{} is a long-studied problem in the field of robotics for various tasks ranging from control~\cite{garcia2008sensor, khalil2010dexterous}, manipulation~\cite{hu2016development, xia2022review}, localization, and navigation~\cite{alatise2020review}, but remains underexplored in RL settings.

To this end, Imitation Learning approaches~\cite{li_see_2022, hao_masked_2023, saxena_mrest_2024} have shown promise in utilizing multisensory data for learning skills, but require experts collecting informative data, thus limiting their usability, robustness and generalization. 
The challenge of data collection becomes particularly pronounced in tasks involving objects with varying or uncertain properties, such as mass or friction. In such cases, substantial and often expensive data acquisition efforts are required to develop a robust policy capable of generalizing effectively across diverse contextual variations~\cite{belkhale_dataquality_2024}. Self-directed exploration in RL, on the other hand, facilitates the development of adaptable strategies that can be generalized across diverse object properties and contexts. These advantages position RL as a compelling approach for learning multisensory contact-rich manipulation~\cite{lee_making_2018,chen_visuo-tactile_2022,liu_masked_2024}. 
Recent advances in multimodal learning \cite{bachmann_multimae_2022, geng_multimodal_2022} have shown the advantages of reconstructing masked or noised inputs to learn expressive cross-modal representations for downstream tasks. Such masking-based self-supervision approaches,  whether at the sensory or embedding level, have also been shown to improve the network's robustness~\cite{skand_simple_2024, liu_learning_2017, radosavovic_real-world_2022}. 

\begin{figure}[t!]
    \centering
    \includegraphics[width=\columnwidth]{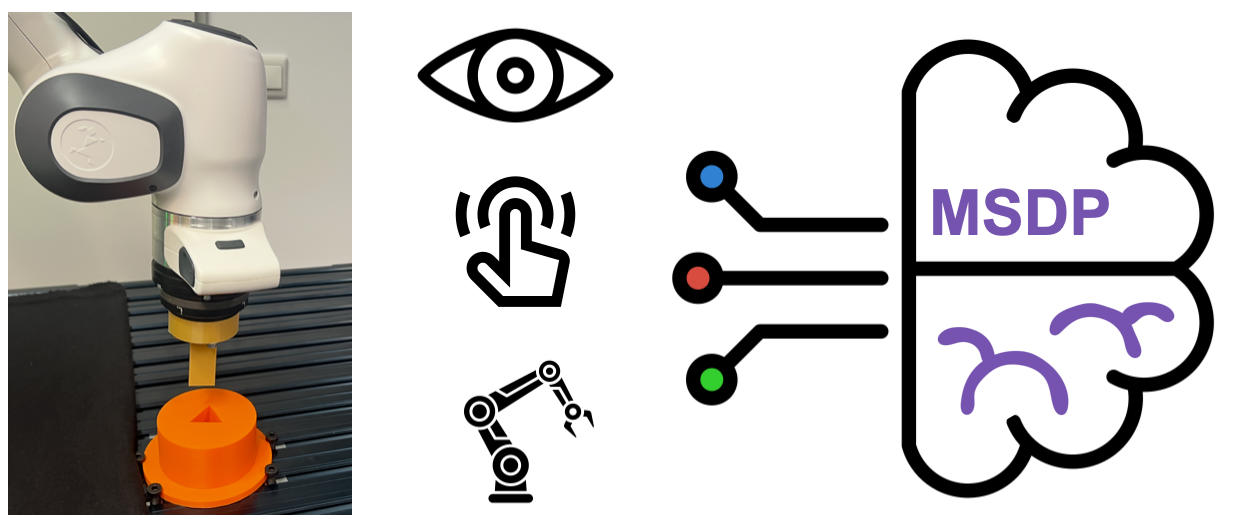}
    \caption{Multisensory Dynamic Pretraining fuses multiple sensors, like human senses, to solve complex contact-rich manipulation tasks.}
    \vspace{-2.0em} 
    \label{fig:teaser}
\end{figure}

In this paper, we propose \emph{\textbf{M}ulti\textbf{S}ensory \textbf{D}ynamic \textbf{P}retraining} (MSDP), a novel RL framework to learn expressive multisensory representations for contact-rich manipulation tasks.
MSDP first learns to extract and fuse sensor features via an offline, pretraining phase based on masked autoencoding and cross-sensor prediction; 
then, an online RL agent leverages the features from the pretrained multisensory encoder through a new combination of cross-attention and pooling mechanisms applied to the critic and actor, respectively. In turn, this design enables RL policies to synergistically fuse multisensory inputs and inherently cope with sensor noise or missing modalities.

Our experimental results demonstrate that MSDP yields effective representations that accelerate RL on a variety of contact-rich manipulation tasks while being robust to sensor noise and changing object dynamics.
The FT-sensor boosts MSDP's performance in two challenging real robot tasks by 14 \% leading to near-optimal performance. 
Notably, the policy is trained directly in the real world on MSDP's multisensory latent representation, without any sim-to-real transfer. We achieve near-optimal performance in only 6,000 online interactions, which takes less than 55 minutes including data collection and pretraining. Beyond these results, MSDP scales naturally to an increasing number of diverse input modalities and only introduces a few learnable parameters in downstream training of RL agents.

To summarize, the key contributions of our work are twofold:
(i) we develop an effective pretraining strategy based on masked autoencoding to form a rich multisensory representation, and 
(ii) we introduce a novel architecture that allows task-specific feature extraction to achieve efficient \diff{(real)}{} robot Reinforcement Learning. Our findings pave the way for \diff{more}{}  adaptive and resilient RL agents to handle multiple input modalities to master complex contact-rich manipulation tasks.

\section{Related Work}
\label{sec-relwork}

\subsection{Reinforcement Learning for Contact-Rich Manipulation}

Tasks where a robotic manipulator has to interact with its environment, either directly or indirectly via a tool, require a good understanding of the interaction forces that shape the task at hand. Solving contact-rich tasks often relies on an accurate estimate of force and dynamics. When such estimates are available, classical control approaches can be adapted to solve the task~\cite{whitney1987historical,whitney1982quasi}. To better handle unseen or difficult contact dynamics in unstructured environments, RL presents itself as an ideal candidate that can learn directly via interactions~\cite{nuttin1997learning}, without accurate state estimation~\cite{levine2016end}. \diff{In contact-rich manipulation, v}{V}arious works have employed RL to learn contact-\diff{based}{rich manipulation} policies on a variety of tasks~\cite{elguea2023review, suomalainen2022survey,liu2021deep}.

\subsection{Multimodal Self-Supervision for Robotics} 

Lee et al. \cite{lee_making_2018} learns a multimodal representation from vision, force torque, and proprioception via MLP-fusion and multiple self-supervised objectives. The frozen representation leads to a robust representation for RL to solve multiple peg insertion tasks. \cite{chen_visuo-tactile_2022} extended the Vision Transformer~\cite{dosovitskiy_image_2020} with a force torque sensor to solve a variety of contact-rich tasks. They additionally incorporate the self-supervised objectives from~\cite{lee_making_2018} to shape a representation using SLAC~\cite{lee_stochastic_2020}. 
\cite{mejia_hearing_2024} pretrains an audio-encoder~\cite{morgado_audio-visual_2021} to combine vision and audio via a transformer decoder for Imitation Learning. The audio signal from the contact microphone provides rich feedback for various manipulation tasks. To also account for different sensor frequencies~\cite{saxena_mrest_2024} developed a multi-resolution policy based on pretrained Vision Language Models to improve inference time using proprioception and force.

\subsection{Deep Sensor Fusion}  

A robust representation is essential to integrate and process various sensory inputs to ensure stable and efficient learning in multisensory RL. Previous architectures often focused on straightforward latent fusion approaches by concatenating the various representations for downstream tasks~\cite{lee2020making,li2023see,li2021reinforcement,guzey2023dexterity,han2024learning}. Feng et al. \cite{feng2024play} take this a step further by adding a subgoal-aware weighting for learning the stage-wise importance of difference sensors. Alternatively, contrastive learning approaches~\cite{dave_multimodal_2024,lygerakis_m2curl_2024} emphasize feature alignment, rather than fusion, to learn a shared latent representation between multiple modalities. However, distilling task-relevant features while maintaining sensor-specific information can be challenging. In contrast, inspired by the success of masked token prediction~\cite{bachmann_multimae_2022}, recent works have also explored how masked multisensory pretraining can enhance representation learning for contact-rich manipulation~\cite{sferrazza_power_2023,liu_masked_2024}.   
\section{Preliminaries}
\label{sec:preliminaries} 

We define a multisensorial POMDP as a tuple $\mathcal{M} = \langle \mathcal{S}, \mathcal{O}^{MS},\ \mathcal{A},\ \mathcal{R},\ \mathcal{P},\ \gamma \rangle$  \cite{kurniawati_partially_2021},
where $\mathcal{S}$ is the true state of the environment.
The agent does not have access to $\mathcal{S}$ and observes it through multiple sensor observations, which we specify as Proprioception $ \mathcal{O}^{P} \in \mathbb{R}^{14}$ \diff{}{(joint position and velocities)}, Force Torque (FT) $ \mathcal{O}^{FT} \in \mathbb{R}^{4 \times 6}$ \diff{}{(FT-sensor at robot wrist)} and Vision $ O^{V} \in \mathbb{R}^{64 \times 64 \times 3}$ \diff{}{(external RGB camera)}. The sensors build up the multisensory Observation-space $\mathcal{O}^{MS} = [\mathcal{O}^{P},\ \mathcal{O}^{FT},\ \mathcal{O}^{V}]$. $\mathcal{A}$ is the action space \diff{}{(cartesian control)}, $\mathcal{R}: \mathcal{S} \times \mathcal{A} \times \mathcal{S} \to \mathbb{R}$ is the reward function, $\mathcal{P}: \mathcal{S} \times \mathcal{A} \to \mathcal{S}$ is the transition kernel, and $\gamma \in [0, 1)$ is the discount factor. Our goal is to learn a policy $\pi:\mathcal{O}^{MS} \to \mathcal{A}$ that maximizes the discounted sum of rewards $\mathbb{E}_{\pi} \lbrack \sum_{t}^\infty \gamma^t r(s_t, a_t, s_{t+1}) \rbrack$. 

We aim to extract an expressive, fused representation from all sensors in order to solve contact-rich manipulation tasks with RL. We only provide the current multisensory observation and no history. We use the off-policy algorithm SAC \cite{haarnoja_soft_2019} in simulation and RLPD~\cite{ball_efficient_2023} for efficiency in the real world, however we highlight that our approach is compatible with any actor-critic algorithm.

\section{Multisensory Dynamic Pretraining}
\label{sec:method}

\begin{figure*}[t!]
        \centering
        \vspace{0.5em}
        \includegraphics[width=1.0\textwidth]{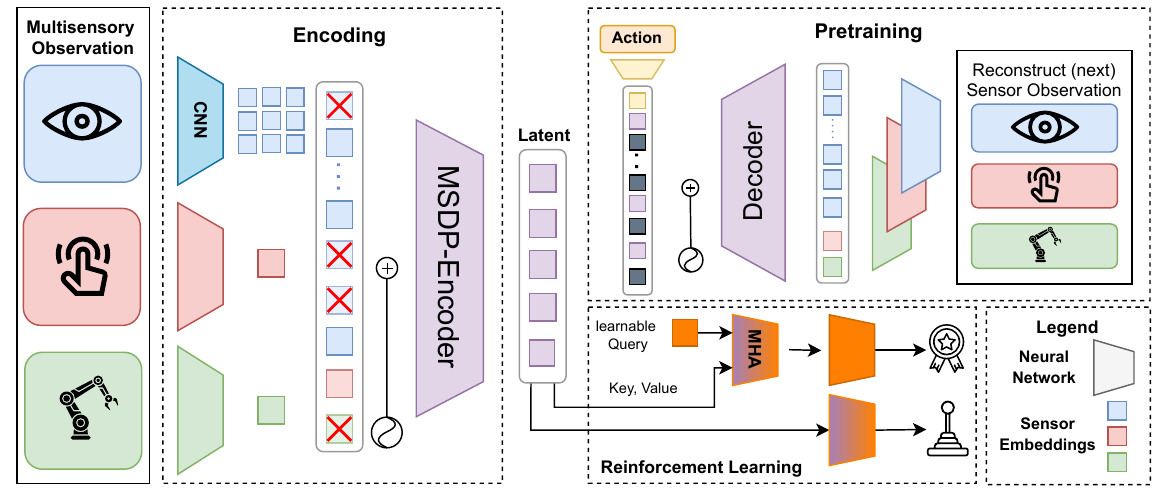}
        
        \caption{The MSDP framework with MSDP-Encoder (left), Pretraining (top right) and downstream RL (bottom right): The current multisensory observation gets projected with a CNN-stem and linear layers to the embedding space. The MSDP-encoder fuses all sensor embeddings to form our expressive multisensory latent representation. The encoder is trained via the decoder and (next) sensor observation reconstruction from a subset of sensor embeddings. This pretraining results in dynamic cross-sensor prediction, shaping and fusing sensor representations. For downstream RL we extract multisensory task-specific features via a single cross-attention layer for the critic and via pooling for the actor. Sensor embeddings are only masked during pretraining. Our Framework offers an expressive and robust multisensory representation for complex contact-rich manipulation tasks in simulation and the real world.}
        \label{fig:overview}
        \vspace{-1.5em}
    \end{figure*}

We present \textbf{M}ulti\textbf{S}ensory  \textbf{D}ynamic \textbf{P}retraining (MSDP), a novel framework for representation learning that builds upon Masked Autoencoders \cite{he_masked_2021}, tailored to enhance contact-rich robot reinforcement learning tasks that require perception through multiple sensor modalities.
MSDP introduces a modular architecture to seamlessly handle diverse input sensors, and a multisensory masking scheme to promote rich cross-modal representations, i.e., to retain knowledge about the task and the environment dynamics even in the absence or perturbation of one or more modalities.

Furthermore, we shed light on the different ways to map pretrained transformer embeddings to input states for downstream RL tasks, referred to as \emph{latent bridging}, a problem often overlooked in practice by the community. 
In this context, we propose employing a simple cross-attention layer to obtain expressive task-specific features from frozen multisensory embeddings for the critic representation, which is also referred to as attention pooling~\cite{chen_context_2023}. The actor on the other hand, receives a special pooling of the sensor embeddings to ensure a stable representation. Both mechanisms offer a fixed low-dimensional observation for the downstream RL agent.

Overall, our framework decouples representation learning and downstream RL to ensure rich feature extraction and fusion, while offering stable, robust and compact downstream representations. Notably, our architecture is scalable w.r.t. the number and types of input modalities and also supports pretraining using additional sensors, which may not be available during policy training. As a result, our pretraining phase on limited offline data leads to rich representations that may be directly used to solve complex contact-rich RL tasks from high-dimensional input (including images) in less than $500k$ environment steps, whereas in the real world only $6k$ online samples are needed.

\vspace{-1.0em}
\subsection{MSDP Architecture}
\label{subsec:maearchitecture}
Given the inputs from multiple sensor modalities, such as vision, force torque readings and proprioception, we encode each modality using a separate network referred to as \emph{sensor encoders}. Particularly for the vision input, we use a \diff{}{4 layer} CNN-stem similar to~\cite{seo_masked_2023}, which allows us to mask at embedding- rather than pixel-level, as masked object positions cannot be retrieved in pixel-space. The CNN-stem introduces redundancy as vision embeddings have overlapping receptive fields and, most importantly, stabilizes training~\cite{xiao_early_2021}. Furthermore, compared to a patchify-stem, it reduces the burden of the multisensory encoder to extract vision features, while fusing with other modalities. We use a linear projection to encode force torque readings and proprioception \diff{}{to the 128-dim embedding space} given their low dimensionalities.

Subsequently, we incorporate position encoding for the embedded sensor features to contextualize the position and modality of each embedding. Rather than using fixed positional/modality encoding~\cite{vaswani_attention_2017}
we use learnable embeddings to encode both together. Once we obtain embedding representations from all sensors, we randomly mask out a subset \diff{}{(70 \%)} of sensor tokens and feed the remaining to our \diff{}{2 layer} transformer encoder \diff{}{with 4 attention heads, pre-normalization and a mlp ratio of 2}\diff{, resulting in multisensory latent embeddings}{}. The attention mechanism of the encoder leads to dynamic multisensory fusion promoted by the representation objective described in Section~\ref{subsec:pretraining}. A learnable mask embedding is added to the randomly masked out embedding positions, next to the multisensory embeddings from the encoder, to generate the original number of embeddings. We reapply our positional/modality encoding before feeding all embeddings to the decoder. During decoding, embeddings exchange information, especially the mask embedding, before being fed to separate decoder heads to fulfill the representation learning objective. We use a shared linear projection for all vision embeddings to reconstruct their corresponding patch. 

\vspace{-0.5em}
\subsection{Representation Learning}
\label{subsec:pretraining}

Our representation learning objective is based on multimodal masked autoencoding~\cite{he_masked_2021,bachmann_multimae_2022, geng_multimodal_2022}. The objective is to either reconstruct the current $\mathbf{O}^{MS}_{t}$ or next observation $\mathbf{O}^{MS}_{t+1}$ from a random subset of sensor embeddings. Vision has a high number of embeddings and needs to extract information about other, potentially masked, sensors e.g. identifying contact to estimate force. Non-vision sensors on the other hand, are beneficial to reconstruct the vision observation as e.g. proprioception defines the robot position. Vision as a global sensing modality never gets fully masked out, whereas low-dimensional sensors are not available when masked. This representation objective results in cross-sensor prediction thus leading to fusion of all modalities. Furthermore, the decoder is conditioned on action in the prediction objective and needs to extract dynamic action-related features important for downstream RL. We denote our method \textbf{MSDP-P} (Prediction) wherein we reconstruct the next observation $||\mathbf{O}^{MS}_{t+1} - \Phi(\mathbf{O}^{MS}_{t}, \mathbf{A}_t)||^2$ and \textbf{MSDP-R} (Reconstruction) where we reconstruct the current observation $||\mathbf{O}^{MS}_{t} - \Phi(\mathbf{O}^{MS}_t)||^2$. $\Phi(.)$ denotes the network prediction conditioned on the current observation $\mathbf{O}^{MS}_t$ and action $\mathbf{A}_t$. $||\cdot||^2$ denotes the Mean Squared Error. Sensor embeddings are only masked during pretraining.

\vspace{-0.5em}
\subsection{Policy Learning}
\label{subsec:policylearning}
From the pretrained encoder, we receive expressive multisensory embeddings given the available sensors. Extracting a compact representation from those embeddings for downstream task-solving is a crucial and often under-discussed design choice. We name the mapping between embeddings and a compact representation \emph{latent bridging}, which has a considerable impact on performance. To address this, we depart from works that naively extract the "CLS"-token from the high-dimensional embeddings, analogous to the Vision Transformer~\cite{radosavovic_real-world_2022, dosovitskiy_image_2020, xiao_masked_2022, majumdar_where_2024} and, instead, propose an asymmetric \emph{latent bridging} strategy between actor and critic to obtain a compact representation. The critic uses a single cross-attention layer with a learnable query and the multisensory embeddings from the MSDP encoder as keys and values (see Figure~\ref{fig:overview}). It offers dynamic task-specific feature extraction (e.g. object positions, robot state, contact) over the task-solving process. Fine-grained understanding of the environment leads to faster convergence compared to a global representation. The policy, on the other hand, does not profit from a cross-attention layer as it may destabilize training. Instead, we mean pool all embeddings originating from the vision sensor before pooling all embeddings, to account for the uneven number of sensor embeddings. Pooling sensor tokens results in a stable and parameter-free latent bridging similar to~\cite{sferrazza_power_2023, seo_masked_2023}. This assymmetric representation between actor and critic, follows \cite{garcin_studying_2025}, where the actor benefits from a stable representation over task stages and the critic from a dynamic-specific representation. The cross-attention layer is trained by the critic, resulting in task-specific feature extraction. We train with the off-policy algorithm SAC in simulation and RLPD in the real world, where we incorporate offline data for pretraining in our replay buffer. \diff{}{For pretraining and policy training we use a batch size of 64 and a learning rate of $3 \times 10^{-4}$.} Our approach is working with any actor-critic RL algorithm.

\vspace{-0.5em}
\section{Experiments and Results}
\label{sec:experiments}
In this section, we present our experimental setting, three competitive baselines, training details and multisensory environments. Furthermore, we ablate the impact of sensor settings and latent bridging mechanisms. Finally, we investigate how pretraining with multiple sensors can enrich the vision representation for downstream RL and showcase our task-specific feature extraction.

\subsection{Baselines and Training Details}
\label{sec:baselines}

We compare our methods against one transformer and two non transformer-based baselines. The latter models extract sensor-specific features with a CNN for vision and an MLP for proprioception and force torque readings. The \textbf{Concat} model fuses the concatenated features with a 2-layer MLP to form the multisensory latent representation~\cite{lee_making_2018}. The \textbf{PoE} model generates separate means and variances for each sensor and fuses them with a product of experts approach. \textbf{VTT}~\cite{chen_visuo-tactile_2022} fuses all sensors via a transformer-encoder and compresses the features via multiple linear layers to a compact latent. As the baseline models introduce a bottleneck, we neglect masking and pretrain the encoders with observation reconstruction to form the latent. All models use a similar decoder architecture for the representation objectives, where proprioception and force torque signals are reconstructed using separate two-layer MLPs and vision via a deconvolutional CNN.

We collect 30,000 random samples from the environment and train each model with its respective representation learning objective (see Section~\ref{subsec:pretraining}) for 30,000 update steps. After pretraining, we freeze the encoder and train the downstream task for 100 epochs. Each epoch consists of 5,000 interaction steps, totaling 500,000 RL update steps. We report the mean and 95 \% confidence interval over 12 runs.

\vspace{-0.5em}
\subsection{Multisensory Environments}

\begin{figure}[t]
    \centering
    \begin{subfigure}[b]{0.49\columnwidth}
        \includegraphics[width=\columnwidth]{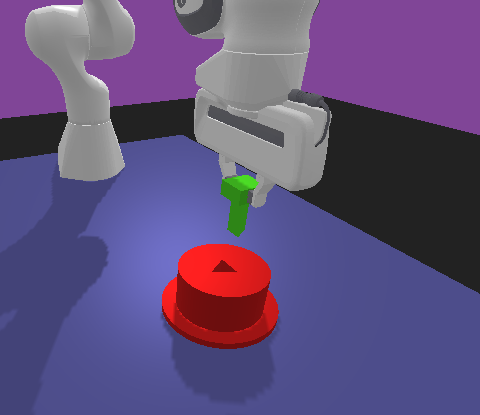}
        \caption{Peg Insertion}
    \end{subfigure}
    \hfill
    \begin{subfigure}[b]{0.49\columnwidth}
        \includegraphics[width=\columnwidth]{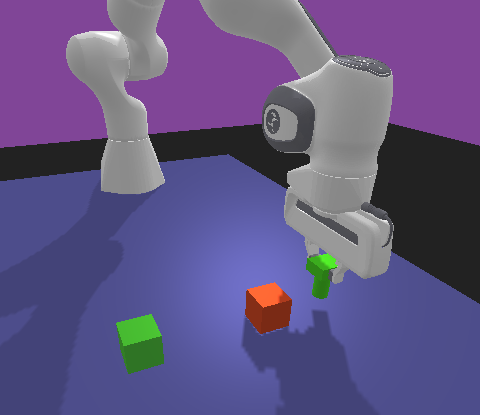}
        \caption{Push Cube}
    \end{subfigure}
    
    \vspace{0.1cm} 
    
    \begin{subfigure}[b]{0.49\columnwidth}
        \includegraphics[width=\columnwidth]{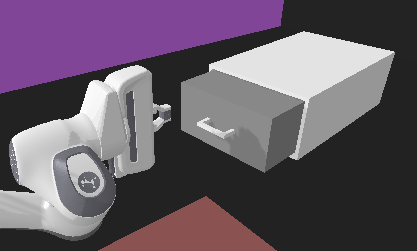}
        \caption{Close Drawer Gently}
    \end{subfigure}
    \hfill
    \begin{subfigure}[b]{0.49\columnwidth}
        \includegraphics[trim={0 0 0 1.6cm},clip,width=\columnwidth]{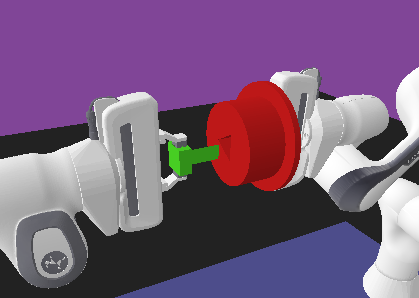}
        \caption{Dual Arm Peg Insertion}
    \end{subfigure}
    \caption{Multisensory contact-rich robot environments}
    \label{fig:environments}
        \vspace{-1.5em}
\end{figure}

\begin{figure*}[t!]
        \centering
        \includegraphics[width=1.0\textwidth]{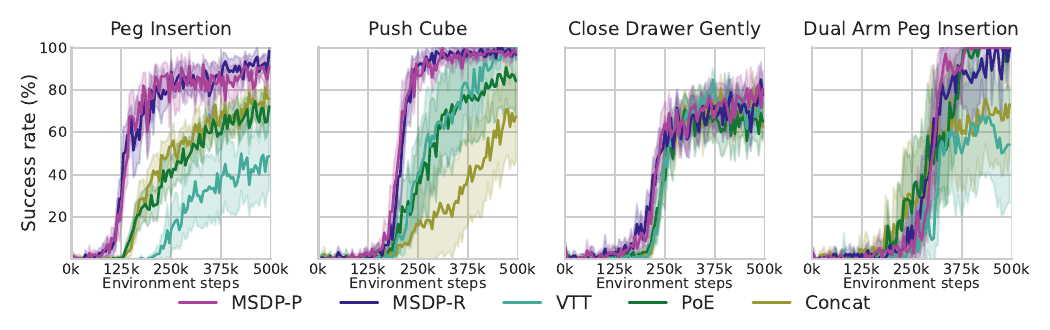}
         \vspace{-1.5em}
        \caption{Performance comparison between MSDP-P and MSDP-R to the baselines in Peg Insertion, Push Cube, Close Drawer Gently and Dual Arm Peg Insertion.
        Our method significantly accelerates RL training and achieves the highest final success rate across all tasks.}
        \label{fig:all_success}
        \vspace{-1.0em}
\end{figure*}

Our multisensory environments are based on panda-gym \cite{gallouedec_panda-gym_2021} and Pybullet \cite{coumans_pybullet_2016}. The tasks Peg Insertion, Push Cube, Close Drawer Gently and Dual Arm Peg Insertion are challenging contact-rich manipulation tasks with sensor noise, varying dynamics and number of sensors. Each task has a dense reward function and comes with multisensory observations consisting of proprioception\diff{(joint position and velocities)}{}, vision \diff{from an external RGB camera}{} and \diff{a}{} force torque \diff{sensor located on the wrist of the robot}{\ readings}. The observation of the Dual Arm environment has an additional FT-sensor and proprioception of the second robot arm. The action space consists of the 3-dimensional endeffector displacement. Task details are provided below and an overview of the environments can be found in Figure~\ref{fig:environments}.

\textbf{Peg Insertion:} Insert the triangular peg into the hole. Next to the position, the z-orientation of the peg needs to match the hole, leading to a 4-dimensional action space and complex insertion dynamics. We randomize the robot initial position, the hole position and orientation. Furthermore, we add gaussian noise to the vision observation to mimic sensor noise and encourage the usage of the FT-sensor.

\textbf{Push Cube:} Push a cube to a target location, with a round peg held by the endeffector. Robot and cube initial position are randomized. Additionally, we vary the cube's mass and center of mass, leading to changing object dynamics. 

\textbf{Close Drawer Gently:} Instead of fully closing a drawer, the task is successful when the drawer is nearly closed, while having a minimal velocity. It requires fine-grained control to be solved, compared to standard close drawer formulations, where undesirable, forceful closing behavior may occur. We randomize the position and orientation of the drawer and the friction of its prismatic joint.

\textbf{Dual Arm Peg Insertion:} extends the Peg Insertion task to a bi-manual setting. We randomize both base positions of the robot and use a similar reward function. Both robot arms need to align the peg and hole positions to solve the task.

\subsection{Simulation Experiments}
\label{subsec:sim_experiments}

Figure~\ref{fig:all_success} shows the performance of our MSDP-P and MSDP-R models compared to the baselines. We obtain superior performance in Peg Insertion and Push Cube, while also outperforming in Close Drawer Gently and Dual Arm Peg Insertion, indicating an expressive representation for contact-rich tasks. Especially in Peg Insertion, our sensor fusion provides fine-grained features to solve the task around 80 \% of the time after only $200,000$ environment steps. The baselines struggle to combine sensor knowledge to consistently insert the peg. Baseline models need an extended amount of training in Push Cube to be able to account for varying object dynamics, like mass and center of mass of the cube, while our representations allow quick adaptation to those changes, leading to a steep learning curve and optimal performance. The Close Drawer Gently task offers no clear results, mainly due to simulation bottlenecks. We observed sudden changes in the drawer during contact that limit the agent's ability to close it smoothly. Still, our method can learn the task faster with better final performance. For the Dual Arm Peg Insertion task additional sensor encoders are used to map the proprioceptive state of the second robot and its force torque reading to the respective latent. Next to PoE we achieve optimal performance over all runs. Our embedding-based latent would allow for more advanced mapping to actor/critic, which we omit for a fair comparison.

\begin{figure}[h!]
    \vspace{-0.5em}
    \centering
    \includegraphics[width=0.9\columnwidth]{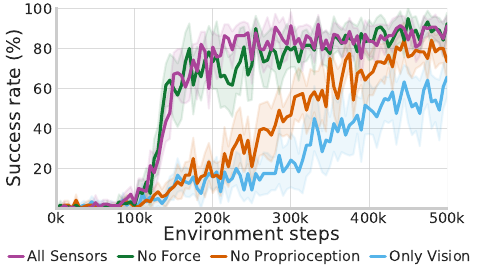}
    \caption{Peg Insertion Sensor Ablation: Proprioception is crucial to identify the precise peg pose under vision noise, while the force torque sensor allows for precise exploration around the hole, resulting in consistent insertions. Vision only is not able to achieve a high success rate.}
        \vspace{-0.5em}  
    \label{fig:peginsert_sensor_ablation}
\end{figure}

\begin{figure}[t!]
    \vspace{0.5em}
    \centering
    \includegraphics[width=0.92\columnwidth]{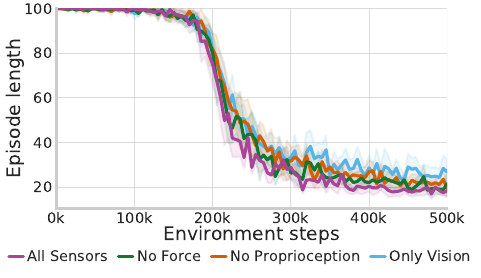}
    \vspace{-0.5em}
    \caption{Push Cube Sensor Ablation: Best performance is achieved using all sensors as the agent is able to maintain contact with the cube in order to push it fast to the goal.}
    \vspace{-1.5em}
    \label{fig:pushcube_sensor_ablation}
\end{figure}

\textbf{Sensor Ablation:} To evaluate the role of individual sensors, we conduct sensor ablation studies on Peg Insertion and Push Cube using the MSDP-P model. Figure~\ref{fig:peginsert_sensor_ablation} compares the success rate in various sensor settings, highlighting the performance gains compared to a vision only approach. Proprioception is essential to identify the peg pose under vision noise, while the force torque sensor guides exploration and allows for consistent insertions, indicated by the lower confidence interval.
Figure~\ref{fig:pushcube_sensor_ablation} shows the episode lengths in the Push Cube task as all sensor combinations achieve near-optimal performance. Vision is sufficient to locate the cube and solve the task. The usage of all sensor allows the agent to make and keep contact with the cube to solve the task in an efficient manner, indicated by the fast completion time (30 \% faster compared to vision only). In addition to the performance gains, we observe smoother object interactions with access to the force torque modality. Most prominent in the Peg Insertion task, where the agent without having access to the FT-sensor asserts high forces on the hole, indicating unwanted exploration actions which may damage the robot or environment in the real world.

\textbf{Latent Bridging:}
The \emph{latent bridging} mechanism to obtain a compact representation from multisensory embeddings, described in Section~\ref{subsec:policylearning}, can have a substantial impact on performance.
We compare our cross-attention extraction with common approaches. \textit{CLS} is using the CLS embedding of the encoder, commonly used in the Vision Transformer~\cite{dosovitskiy_image_2020} or Imitation Learning~\cite{saxena_mrest_2024}. In \textit{Pooling} we mean-pool all sensor embeddings similar to~\cite{sferrazza_power_2023, seo_masked_2023}. To account for the different number of embeddings between sensors, we first take the mean of all vision embeddings. To avoid possible dilution \textit{Cat} provides the Concatenation of sensor embeddings, where we again pool the vision embeddings.  Common latent bridging mechanisms do not introduce new learnable parameters, while our cross-attention only adds minimal parameters to extract task-specific features.

\begin{figure}[h] 
    \centering
     \vspace{-0.5em}
    \includegraphics[width=0.9\columnwidth]{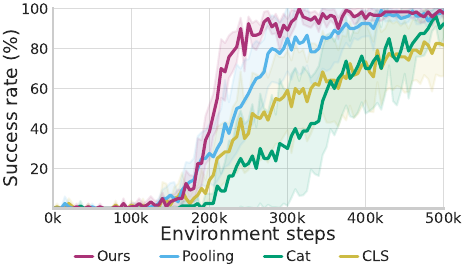}
    \vspace{-0.5em}
    \caption{Push Cube latent bridging ablation: Our mechanism extracts fine-grained multisensory features resulting in efficient policy training.}
    \vspace{-0.5em}
    \label{fig:cube_latents}
\end{figure}
Figure~\ref{fig:cube_latents} shows different latent bridging mechanisms on our MSDP-P model in the Push Cube task. Our approach successfully extracts task-specific features over an episode in an efficient manner.  Our mechanism can extract more fine-grained vision features and combine them with other sensor information, resulting in superior performance. The \textit{CLS} embedding doesn't contain rich features for downstream task learning, while \textit{Pooling} takes longer to solve the task consistently, as the agent needs to extract environment details from a potentially diluted representation.

\begin{figure}[t]
    \centering
    \vspace{0.5em}
    \includegraphics[width=0.94\columnwidth]{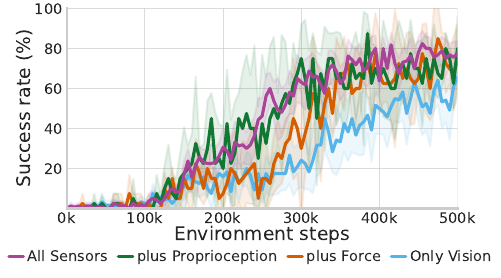}
    \vspace{-0.4em}
    \caption{Pretraining the representation for Peg Insertion with multiple sensors improves performance, when only vision is available during policy training.}
    \vspace{-1.5em}
    \label{fig:peg_res_rlonlyv}
\end{figure}

\textbf{Enriched Representation for vision-based RL:} Our MSDP transformer encoder allows for varying input length, offering us the option to pretrain the representation on more sensors and only use a subset, for downstream task learning. We pretrain our MSDP-P model with different sensor configurations and only provide the vision modality during task solving. Figure~\ref{fig:peg_res_rlonlyv} shows how different pretraining settings can enhance vision-based RL in the Peg Insertion task. The encoder stores information about proprioception and force torque readings promoted by the masked autoencoding pretraining objective, thus enriching the vision representation, leading to increased performance. 

\begin{figure}[h]
    \centering
    \vspace{-1.0em}
    \begin{subfigure}[t]{0.24\columnwidth}
        \centering
        \includegraphics[width=\linewidth]{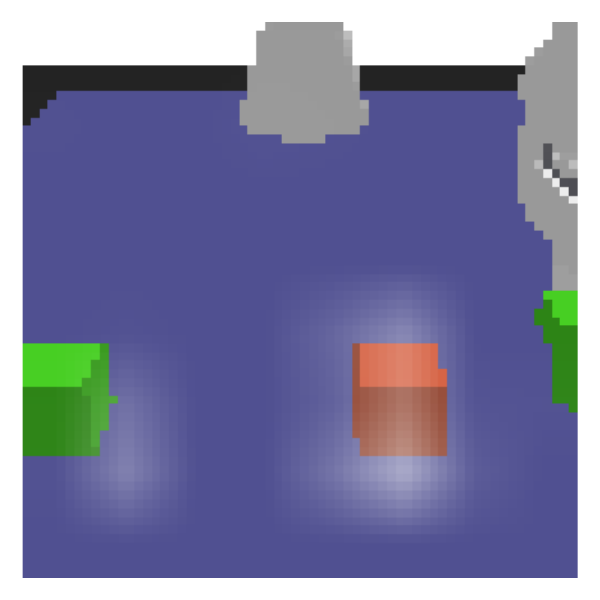}
        \label{fig:sub1}
    \end{subfigure}
    \hfill
    \begin{subfigure}[t]{0.24\columnwidth}
        \centering
        \includegraphics[width=\linewidth]{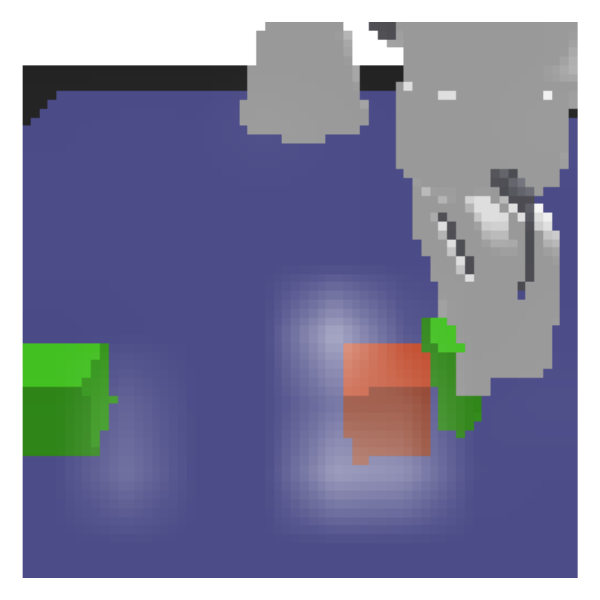}
        \label{fig:sub2}
    \end{subfigure}
    \hfill
    \begin{subfigure}[t]{0.24\columnwidth}
        \centering
        \includegraphics[width=\linewidth]{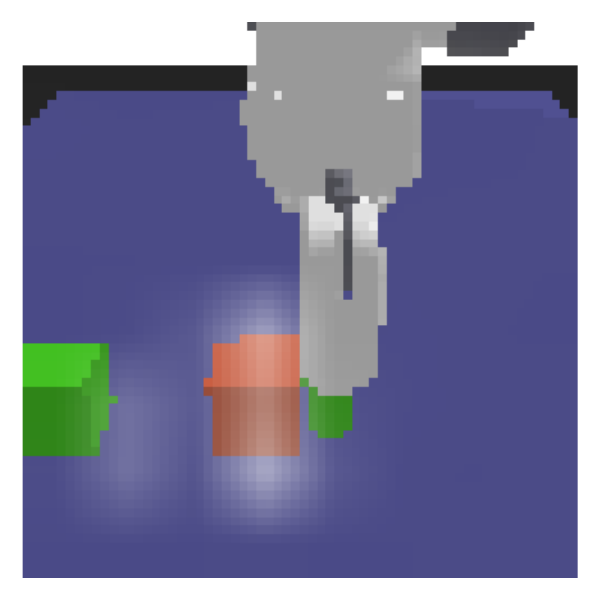}
        \label{fig:sub3}
    \end{subfigure}
    \hfill
    \begin{subfigure}[t]{0.24\columnwidth}
        \centering
        \includegraphics[width=\linewidth]{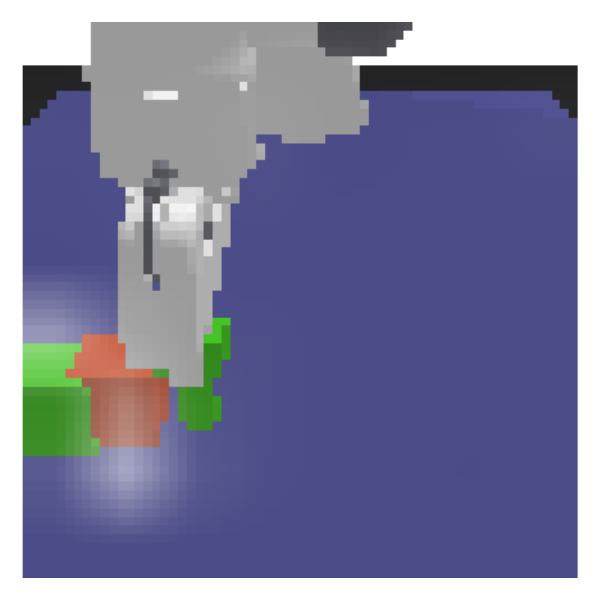}
        \label{fig:sub4}
    \end{subfigure}

    \vspace{-1.5em}
    \caption{Critic's cross-attention maps in the Push Cube task. The critic is focusing on the multisensory embeddings originated at the cube position, indicating task-specific feature extraction.}
    \vspace{-1.5em}
    \label{fig:CrossQ_maps}

\end{figure}

\textbf{Task-specific features extraction:} The cross-attention map in figure~\ref{fig:CrossQ_maps} highlights how the critic focuses on task-relevant features e.g. cube and target position. We want to stress that all embeddings from the pretrained encoder are multisensorial, even if we can assign their origin from a specific sensor. As shown in the previous paragraph, additional sensor features are encoded in the vision modality.

\vspace{-0.5em}
\subsection{Real World Experiments}
\label{subsec:real_world}

\begin{figure*}[t!]
    \centering
    \vspace{1em}
    \begin{subfigure}[t]{0.48\textwidth}
        \begin{subfigure}[t]{0.24\columnwidth}
            \centering
            \includegraphics[width=\linewidth]{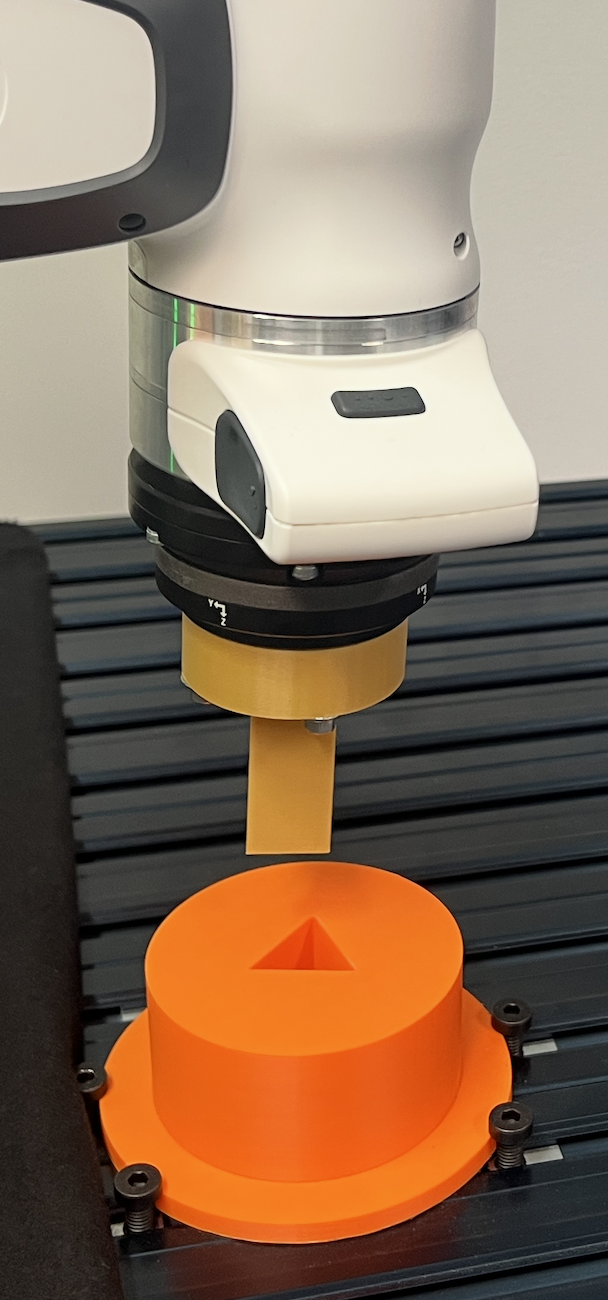}
            \label{fig:peg_image}
        \end{subfigure}
        \hfill
        \begin{subfigure}[t]{0.74\columnwidth}
            \centering
            \includegraphics[width=\linewidth]{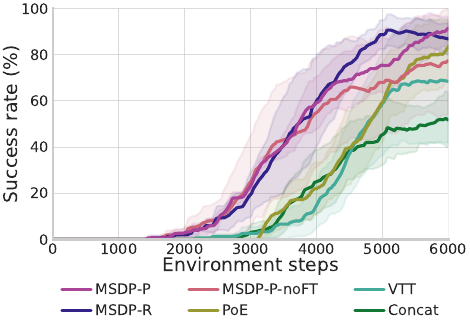}
            \label{fig:peg_curves}
        \end{subfigure}
        \vspace{-1.0em}
        \caption{Real World Peg Insertion.}
        \label{fig:real_peg}
    \end{subfigure}
    \begin{subfigure}[t]{0.48\textwidth}
        \begin{subfigure}[t]{0.245\columnwidth}
            \centering
            \includegraphics[width=\linewidth]{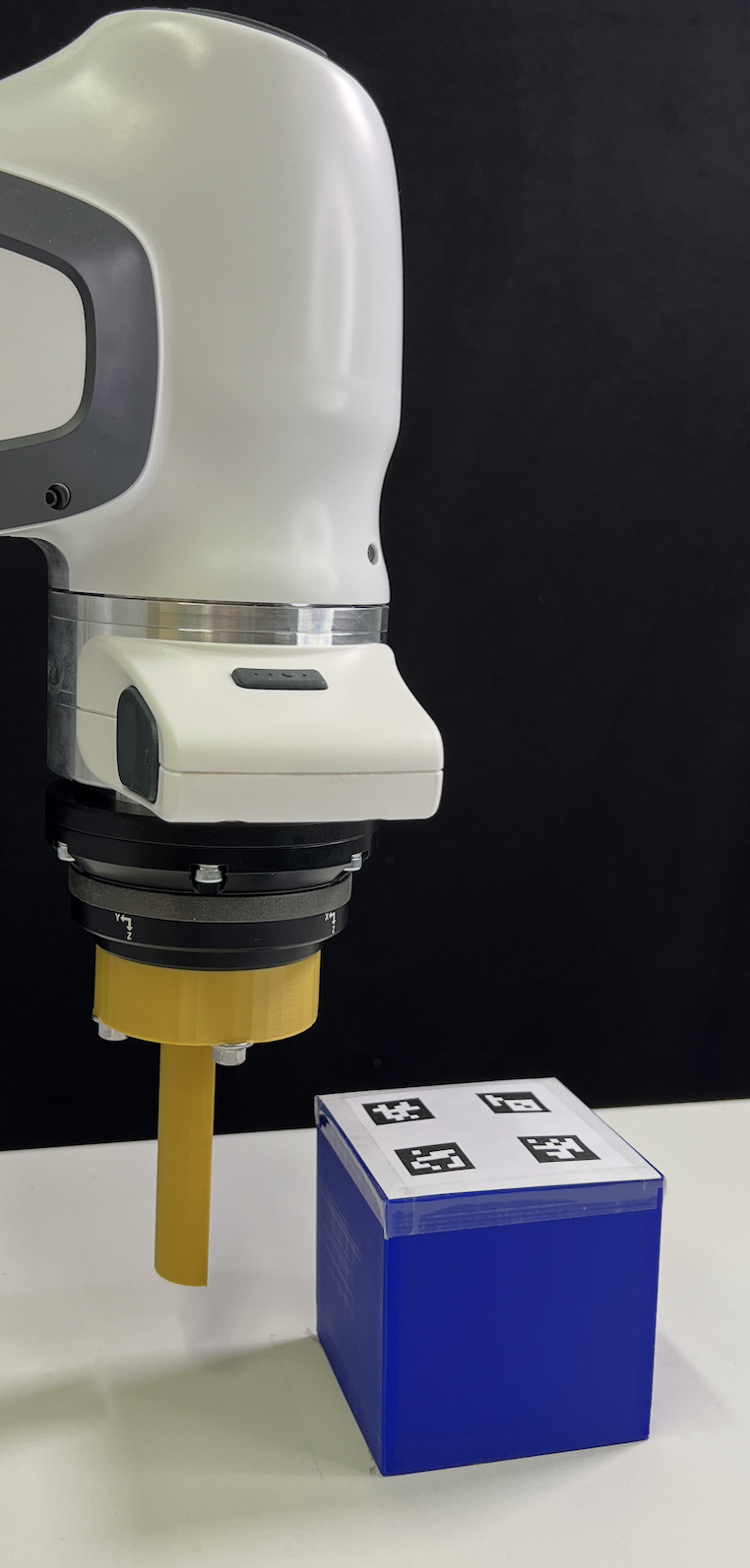}
            \label{fig:push_image}
        \end{subfigure}
        \hfill
        \begin{subfigure}[t]{0.735\columnwidth}
            \centering
            \includegraphics[width=\linewidth]{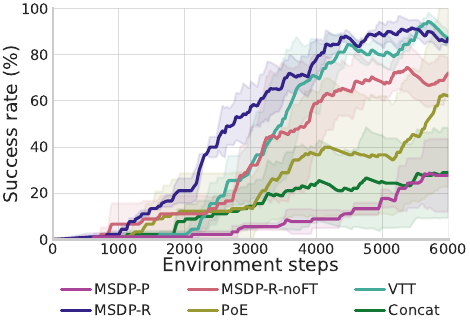}
            \label{fig:push_curves}
        \end{subfigure}
        \vspace{-1.0em}
        \caption{Real World Push Cube.}
        \label{fig:real_pushcube}
    \end{subfigure}
    \caption{Real world setup and experimental results.
    Our MSDP framework enables training RL policies directly in the real world,
    with first successful episodes after only 2,000 / 1,000 online interactions, outperforming various baselines. Task success is detected via endeffector position or the aruco marker on the cube. Force torque readings are essential to consistently push the cube to the goal or to insert the peg under occlusion improving task success by 14 \% (cf. MSDP-noFT). Policies are learned directly on the pretrained multisensory representation, without any sim-to-real transfer, with only 6,000 online interactions.}
    \vspace{-1.5em}
\end{figure*}

We conduct the Peg Insertion and Push Cube tasks in the real world using a Franka robot arm with a wrist-mounted FT-sensor and a custom endeffector. The observation space consists of endeffector position and velocity as proprioception, four force torque readings and an downsampled 64 by 64 RGB image from the 3rd person RealSense camera for both tasks. Peg and hole are both 3d printed. The robot needs to align the orientation of the triangular peg and place it precisely to fully insert the peg and complete the task. In Push Cube the agent needs to push the 7.5 cm block to the 15 cm away goal location. Successful episodes are detected with the endeffector position in Peg Insertion and an aruco marker and a second camera in the Push Cube task. The starting position of the robot is randomized and we use a sparse reward of +1 upon success in both tasks. An overview of the setup and results can be found in Figure~\ref{fig:real_peg} and \ref{fig:real_pushcube}.

We build upon the SERL-package \cite{luo2024serl} and use a cartesian impedance controller for safe interaction. The multisensory encoder is pretrained with data from 20 demonstrations and \textasciitilde 2,000 samples of play data, resulting in a total of \textasciitilde 3,000 samples. Downstream tasks are trained directly on the real system for only 6,000 environment interactions with the proposed RLPD~\cite{ball_efficient_2023} algorithm from SERL. We pretrain for 6,000 steps in Peg Insertion, updating critic and actor twice per online sample. For Push Cube, due to higher vision variance, we use 10,000 update steps and apply four critic and one actor update step per online interaction to obtain meaningful representations and efficient training. The complete training pipeline, including data collection, pretraining, and online RL, learns the final policy in less than 55 minutes for both tasks. \diff{}{Compared to prior real-world multisensory RL methods~\cite{lee_making_2018, chen_visuo-tactile_2022}, our approach requires only a fraction of the data (3 \% pretraining data in Peg Insertion) and training time (< 20 \%) to learn the final policy.}

Figures~\ref{fig:real_peg} and \ref{fig:real_pushcube} show results for Peg Insertion (5 runs) and Push Cube (3 runs), respectively. Both our pretraining objectives achieve superior performance in Peg Insertion, whereas baselines suffer from vision noise and occlusion.
Particularly, we notice that our synergistic use of force torque readings enables consistent insertion behavior, while reducing the asserted force between the peg and the hole during exploration.
On the other hand, the Push Cube task introduces several challenges as the agent needs to make and maintain contact with the cube without pushing it out of the workspace.
Here, MSDP-R and VTT are able to obtain a high final success rate (see Figure~\ref{fig:real_pushcube}), while MSDP-P is not able to extract a suitable representation. We attribute this shortcoming to the complexity of learning forward dynamics in vision-demanding tasks on limited real world data and lack of observation histories.
Overall, we highlight that our expressive multisensory representation, in combination with task-specific feature extraction via latent bridging, is able to obtain first success with less than 2,000 online interactions. 
%
FT-sensor usage particularly improves the performance of MSDP by 14 \% in both real-world tasks, underscoring the significance of multi-sensor integration in contact-rich manipulation.

\diff{Using online available datasets for pretraining has been omitted in this work, due to limited multisensory datasets and a large domain gap.}{}

\begin{figure}[t]
    \centering
    \begin{subfigure}[t]{0.24\columnwidth}
        \centering
        \includegraphics[width=\linewidth]{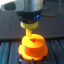}
        \label{fig:sub1}
    \end{subfigure}
    \hfill
    \begin{subfigure}[t]{0.24\columnwidth}
        \centering
        \includegraphics[width=\linewidth]{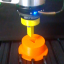}
        \label{fig:sub2}
    \end{subfigure}
    \hfill
    \begin{subfigure}[t]{0.24\columnwidth}
        \centering
        \includegraphics[width=\linewidth]{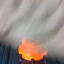}
        \label{fig:sub3}
    \end{subfigure}
    \hfill
    \begin{subfigure}[t]{0.24\columnwidth}
        \centering
        \includegraphics[width=\linewidth]{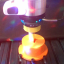}
        \label{fig:sub4}
    \end{subfigure}

    \vspace{-1.0em}
    \caption{\diff{}{Visual disturbances of the robustness evaluation. From left to right: back light, front light, occlusion, and disco lights. MSDP achieves a high success rate across all disturbances, which have not been seen during training.}}
    \vspace{-1.5em}
    \label{fig:vision_robustness}

\end{figure}

\diff{}{\textbf{Robustness Evaluation:} We evaluate the final policy of MSDP-P in the Peg Insertion task under various disturbances to showcase the robustness of our pretrained multisensory encoder and policy. We evaluate each change that hasn't been observed during training for 20 trials. Trained on $K_c=2000$ cartesian-stiffness the policy achieves 90 \% success rate with decreased ($K_c=1500$) and 100 \% with increased cartesian-stiffness ($K_c=2500$). MSDP shows remarkable robustness against changed light settings, e.g., back light (100 \%), front light (100 \%), disco lights (100 \%) and visual occlusion (partly blocked camera view, 95 \%) and external forces (80 \%). All visual disturbances are shown in Figure~\ref{fig:vision_robustness}.}

\diff{}{This work assumes the presence of all sensor modalities at the start of training to obtain an expressive representation from limited multisensory data. The usage of MSDP to enhance existing representations seems promising to leverage common available datasets for generalization.}


\section{Conclusion and Future Work}
\label{sec:conclusion}

In this work, we propose MultiSensory Dynamic Pretraining (MSDP), a novel pretraining framework to learn multisensory representations for contact-rich manipulation tasks. Specifically, MSDP learns to reconstruct varying sensory information from a subset of input sensor embeddings, leading to cross-sensor predictions and sensor fusion.
MSDP captures the interplay between different sensor modalities to learn a rich multisensory representation, which in combination with task-specific feature extraction through cross-attention, leads to a superior performance in challenging contact-rich manipulation tasks in simulation and the real world.
Future work may investigate MSDP's capabilities for sensor modalities beyond vision and force-torque, such as tactile or audio\diff{}{, as well as its use for fine-tuning existing representations with multisensory data.}
Finally, we reckon that extending MSDP's pretraining stage to incorporate play data from environments at large scale may yield representations that generalize over new tasks.





\bibliographystyle{IEEEtran}

\bibliography{RAL/MSRL_ral}

@article{belkhale_dataquality_2024,
  title={Data quality in imitation learning},
  author={Belkhale, Suneel and Cui, Yuchen and Sadigh, Dorsa},
  journal={Advances in Neural Information Processing Systems},
  volume={36},
  year={2024}
}

@article{dosovitskiy_image_2020,
	title = {An {Image} is {Worth} 16x16 {Words}: {Transformers} for {Image} {Recognition} at {Scale}},
	volume = {abs/2010.11929},
	url = {https://arxiv.org/abs/2010.11929},
	journal = {CoRR},
	author = {Dosovitskiy, Alexey and Beyer, Lucas and Kolesnikov, Alexander and Weissenborn, Dirk and Zhai, Xiaohua and Unterthiner, Thomas and Dehghani, Mostafa and Minderer, Matthias and Heigold, Georg and Gelly, Sylvain and Uszkoreit, Jakob and Houlsby, Neil},
	year = {2020},
	note = {\_eprint: 2010.11929},
}

@inproceedings{vaswani_attention_2017,
	title = {Attention is {All} you {Need}},
	volume = {30},
	url = {https://proceedings.neurips.cc/paper/2017/file/3f5ee243547dee91fbd053c1c4a845aa-Paper.pdf},
	booktitle = {Advances in {Neural} {Information} {Processing} {Systems}},
	publisher = {Curran Associates, Inc.},
	author = {Vaswani, Ashish and Shazeer, Noam and Parmar, Niki and Uszkoreit, Jakob and Jones, Llion and Gomez, Aidan N and Kaiser, Lukasz and Polosukhin, Illia},
	editor = {Guyon, I. and Luxburg, U. V. and Bengio, S. and Wallach, H. and Fergus, R. and Vishwanathan, S. and Garnett, R.},
	year = {2017},
}

@article{lee_making_2018,
	title = {Making {Sense} of {Vision} and {Touch}: {Self}-{Supervised} {Learning} of {Multimodal} {Representations} for {Contact}-{Rich} {Tasks}},
	volume = {abs/1810.10191},
	url = {http://arxiv.org/abs/1810.10191},
	journal = {CoRR},
	author = {Lee, Michelle A. and Zhu, Yuke and Srinivasan, Krishnan and Shah, Parth and Savarese, Silvio and Fei-Fei, Li and Garg, Animesh and Bohg, Jeannette},
	year = {2018},
	note = {\_eprint: 1810.10191},
}

@misc{chen_visuo-tactile_2022,
	title = {Visuo-{Tactile} {Transformers} for {Manipulation}},
	url = {http://arxiv.org/abs/2210.00121 April 3rd},
	doi = {10.48550/arXiv.2210.00121},
	urldate = {2023-08-01},
	publisher = {arXiv},
	author = {Chen, Yizhou and Sipos, Andrea and Van der Merwe, Mark and Fazeli, Nima},
		year = {2022},
	note = {arXiv:2210.00121 [cs]},
}

@misc{coumans_pybullet_2016,
	title = {{PyBullet}, a {Python} module for physics simulation for games, robotics and machine learning},
	url = {http://pybullet.org},
	author = {Coumans, Erwin and Bai, Yunfei},
	year = {2016},
}

@article{gallouedec_panda-gym_2021,
	title = {panda-gym: {Open}-{Source} {Goal}-{Conditioned} {Environments} for {Robotic} {Learning}},
	journal = {4th Robot Learning Workshop: Self-Supervised and Lifelong Learning at NeurIPS},
	author = {Gallouédec, Quentin and Cazin, Nicolas and Dellandréa, Emmanuel and Chen, Liming},
	year = {2021},
}

@misc{haarnoja_soft_2019,
	title = {Soft {Actor}-{Critic} {Algorithms} and {Applications}},
	author = {Haarnoja, Tuomas and Zhou, Aurick and Hartikainen, Kristian and Tucker, George and Ha, Sehoon and Tan, Jie and Kumar, Vikash and Zhu, Henry and Gupta, Abhishek and Abbeel, Pieter and Levine, Sergey},
	year = {2019},
	note = {\_eprint: 1812.05905},
}

@misc{sferrazza_power_2023,
	title = {The {Power} of the {Senses}: {Generalizable} {Manipulation} from {Vision} and {Touch} through {Masked} {Multimodal} {Learning}},
	shorttitle = {The {Power} of the {Senses}},
	url = {http://arxiv.org/abs/2311.00924},
	doi = {10.48550/arXiv.2311.00924},
	urldate = {2024-02-21},
	publisher = {arXiv},
	author = {Sferrazza, Carmelo and Seo, Younggyo and Liu, Hao and Lee, Youngwoon and Abbeel, Pieter},
		year = {2023},
	note = {arXiv:2311.00924 [cs]},
}

@misc{kurniawati_partially_2021,
	title = {Partially {Observable} {Markov} {Decision} {Processes} ({POMDPs}) and {Robotics}},
	url = {http://arxiv.org/abs/2107.07599},
	doi = {10.48550/arXiv.2107.07599},
	urldate = {2024-05-21},
	publisher = {arXiv},
	author = {Kurniawati, Hanna},
		year = {2021},
	note = {arXiv:2107.07599 [cs]},
}

@inproceedings{li2023see,
  title={See, Hear, and Feel: Smart Sensory Fusion for Robotic Manipulation},
  author={Li, Hao and Zhang, Yizhi and Zhu, Junzhe and Wang, Shaoxiong and Lee, Michelle A and Xu, Huazhe and Adelson, Edward and Fei-Fei, Li and Gao, Ruohan and Wu, Jiajun},
  booktitle={Conference on Robot Learning},
  pages={1368--1378},
  year={2023},
  organization={PMLR}
}

@misc{saxena_mrest_2024,
	title = {{MResT}: {Multi}-{Resolution} {Sensing} for {Real}-{Time} {Control} with {Vision}-{Language} {Models}},
	shorttitle = {{MResT}},
	url = {http://arxiv.org/abs/2401.14502},
	doi = {10.48550/arXiv.2401.14502},
	urldate = {2024-06-11},
	publisher = {arXiv},
	author = {Saxena, Saumya and Sharma, Mohit and Kroemer, Oliver},
		year = {2024},
	note = {arXiv:2401.14502 [cs]},
}

@misc{seo_masked_2023,
	title = {Masked {World} {Models} for {Visual} {Control}},
	url = {http://arxiv.org/abs/2206.14244},
	doi = {10.48550/arXiv.2206.14244},
	urldate = {2024-06-12},
	publisher = {arXiv},
	author = {Seo, Younggyo and Hafner, Danijar and Liu, Hao and Liu, Fangchen and James, Stephen and Lee, Kimin and Abbeel, Pieter},
		year = {2023},
	note = {arXiv:2206.14244 [cs]},
}

@misc{radosavovic_real-world_2022,
	title = {Real-{World} {Robot} {Learning} with {Masked} {Visual} {Pre}-training},
	url = {http://arxiv.org/abs/2210.03109},
	doi = {10.48550/arXiv.2210.03109},
	urldate = {2024-06-20},
	publisher = {arXiv},
	author = {Radosavovic, Ilija and Xiao, Tete and James, Stephen and Abbeel, Pieter and Malik, Jitendra and Darrell, Trevor},
		year = {2022},
	note = {arXiv:2210.03109 [cs]},
}

@misc{geng_multimodal_2022,
	title = {Multimodal {Masked} {Autoencoders} {Learn} {Transferable} {Representations}},
	url = {https://arxiv.org/abs/2205.14204v3},
	language = {en},
	urldate = {2024-09-06},
	journal = {arXiv.org},
	author = {Geng, Xinyang and Liu, Hao and Lee, Lisa and Schuurmans, Dale and Levine, Sergey and Abbeel, Pieter},
	year = {2022},
}

@misc{bachmann_multimae_2022,
	title = {{MultiMAE}: {Multi}-modal {Multi}-task {Masked} {Autoencoders}},
	shorttitle = {{MultiMAE}},
	url = {https://arxiv.org/abs/2204.01678v1},
	language = {en},
	urldate = {2024-09-06},
	journal = {arXiv.org},
	author = {Bachmann, Roman and Mizrahi, David and Atanov, Andrei and Zamir, Amir},
		year = {2022},
}

@inproceedings{skand_simple_2024,
	title = {Simple {Masked} {Training} {Strategies} {Yield} {Control} {Policies} {That} {Are} {Robust} to {Sensor} {Failure}},
	url = {https://openreview.net/forum?id=AsbyZRdqPv},
	language = {en},
	urldate = {2024-09-12},
	author = {Skand, Skand and Pandit, Bikram and Kim, Chanho and Fuxin, Li and Lee, Stefan},
		year = {2024},
}

@misc{xiao_early_2021,
	title = {Early {Convolutions} {Help} {Transformers} {See} {Better}},
	url = {http://arxiv.org/abs/2106.14881},
	doi = {10.48550/arXiv.2106.14881},
	urldate = {2024-09-13},
	publisher = {arXiv},
	author = {Xiao, Tete and Singh, Mannat and Mintun, Eric and Darrell, Trevor and Dollár, Piotr and Girshick, Ross},
		year = {2021},
	note = {arXiv:2106.14881 [cs]},
}

@book{liu_masked_2024,
	title = {Masked {Visual}-{Tactile} {Pre}-training for {Robot} {Manipulation}},
	author = {Liu, Qingtao and Sun, Zhengnan and Cui, Yu and Gaofeng, Li and Ye, Qi and Chen, Jiming},
		year = {2024},
	doi = {10.1109/ICRA57147.2024.10610933},
}

@misc{he_masked_2021,
	title = {Masked {Autoencoders} {Are} {Scalable} {Vision} {Learners}},
	url = {http://arxiv.org/abs/2111.06377},
	doi = {10.48550/arXiv.2111.06377},
	urldate = {2024-09-16},
	publisher = {arXiv},
	author = {He, Kaiming and Chen, Xinlei and Xie, Saining and Li, Yanghao and Dollár, Piotr and Girshick, Ross},
	year = {2021},
	note = {arXiv:2111.06377 [cs]},
}

@misc{mejia_hearing_2024,
	title = {Hearing {Touch}: {Audio}-{Visual} {Pretraining} for {Contact}-{Rich} {Manipulation}},
	shorttitle = {Hearing {Touch}},
	url = {http://arxiv.org/abs/2405.08576},
	doi = {10.48550/arXiv.2405.08576},
	urldate = {2024-09-16},
	publisher = {arXiv},
	author = {Mejia, Jared and Dean, Victoria and Hellebrekers, Tess and Gupta, Abhinav},
		year = {2024},
	note = {arXiv:2405.08576 [cs]},
}

@misc{li_see_2022,
	title = {See, {Hear}, and {Feel}: {Smart} {Sensory} {Fusion} for {Robotic} {Manipulation}},
	shorttitle = {See, {Hear}, and {Feel}},
	url = {http://arxiv.org/abs/2212.03858},
	doi = {10.48550/arXiv.2212.03858},
	urldate = {2024-09-20},
	publisher = {arXiv},
	author = {Li, Hao and Zhang, Yizhi and Zhu, Junzhe and Wang, Shaoxiong and Lee, Michelle A. and Xu, Huazhe and Adelson, Edward and Fei-Fei, Li and Gao, Ruohan and Wu, Jiajun},
		year = {2022},
	note = {arXiv:2212.03858 [cs]},
}

@inproceedings{hao_masked_2023,
	title = {Masked {Imitation} {Learning}: {Discovering} {Environment}-{Invariant} {Modalities} in {Multimodal} {Demonstrations}},
	shorttitle = {Masked {Imitation} {Learning}},
	url = {https://ieeexplore.ieee.org/document/10341728/?arnumber=10341728},
	doi = {10.1109/IROS55552.2023.10341728},
	urldate = {2024-09-20},
	booktitle = {2023 {IEEE}/{RSJ} {International} {Conference} on {Intelligent} {Robots} and {Systems} ({IROS})},
	author = {Hao, Yilun and Wang, Ruinan and Cao, Zhangjie and Wang, Zihan and Cui, Yuchen and Sadigh, Dorsa},
		year = {2023},
	note = {ISSN: 2153-0866},
	pages = {1--7},
}

@article{liu_learning_2017,
	title = {Learning {End}-to-end {Multimodal} {Sensor} {Policies} for {Autonomous} {Navigation}},
	language = {en},
	author = {Liu, Guan-Horng and Siravuru, Avinash and Prabhakar, Sai and Veloso, Manuela and Kantor, George},
	year = {2017},
}

@misc{xiao_masked_2022,
	title = {Masked {Visual} {Pre}-training for {Motor} {Control}},
	url = {http://arxiv.org/abs/2203.06173},
	doi = {10.48550/arXiv.2203.06173},
	urldate = {2024-11-11},
	publisher = {arXiv},
	author = {Xiao, Tete and Radosavovic, Ilija and Darrell, Trevor and Malik, Jitendra},
		year = {2022},
	note = {arXiv:2203.06173},
}

@misc{morgado_audio-visual_2021,
	title = {Audio-{Visual} {Instance} {Discrimination} with {Cross}-{Modal} {Agreement}},
	url = {http://arxiv.org/abs/2004.12943},
	doi = {10.48550/arXiv.2004.12943},
	urldate = {2025-01-25},
	publisher = {arXiv},
	author = {Morgado, Pedro and Vasconcelos, Nuno and Misra, Ishan},
		year = {2021},
	note = {arXiv:2004.12943 [cs]},
}

@misc{lee_stochastic_2020,
	title = {Stochastic {Latent} {Actor}-{Critic}: {Deep} {Reinforcement} {Learning} with a {Latent} {Variable} {Model}},
	shorttitle = {Stochastic {Latent} {Actor}-{Critic}},
	url = {http://arxiv.org/abs/1907.00953},
	doi = {10.48550/arXiv.1907.00953},
	urldate = {2025-01-25},
	publisher = {arXiv},
	author = {Lee, Alex X. and Nagabandi, Anusha and Abbeel, Pieter and Levine, Sergey},
		year = {2020},
	note = {arXiv:1907.00953 [cs]},
}

@misc{majumdar_where_2024,
	title = {Where are we in the search for an {Artificial} {Visual} {Cortex} for {Embodied} {Intelligence}?},
	url = {http://arxiv.org/abs/2303.18240},
	doi = {10.48550/arXiv.2303.18240},
	urldate = {2025-01-13},
	publisher = {arXiv},
	author = {Majumdar, Arjun and Yadav, Karmesh and Arnaud, Sergio and Ma, Yecheng Jason and Chen, Claire and Silwal, Sneha and Jain, Aryan and Berges, Vincent-Pierre and Abbeel, Pieter and Malik, Jitendra and Batra, Dhruv and Lin, Yixin and Maksymets, Oleksandr and Rajeswaran, Aravind and Meier, Franziska},
		year = {2024},
	note = {arXiv:2303.18240 [cs]},
}

@misc{dave_multimodal_2024,
	title = {Multimodal {Visual}-{Tactile} {Representation} {Learning} through {Self}-{Supervised} {Contrastive} {Pre}-{Training}},
	url = {http://arxiv.org/abs/2401.12024},
	doi = {10.48550/arXiv.2401.12024},
	urldate = {2025-01-16},
	publisher = {arXiv},
	author = {Dave, Vedant and Lygerakis, Fotios and Rueckert, Elmar},
		year = {2024},
	note = {arXiv:2401.12024 [cs]},
}

@misc{lygerakis_m2curl_2024,
	title = {{M2CURL}: {Sample}-{Efficient} {Multimodal} {Reinforcement} {Learning} via {Self}-{Supervised} {Representation} {Learning} for {Robotic} {Manipulation}},
	shorttitle = {{M2CURL}},
	url = {http://arxiv.org/abs/2401.17032},
	doi = {10.48550/arXiv.2401.17032},
	urldate = {2025-01-16},
	publisher = {arXiv},
	author = {Lygerakis, Fotios and Dave, Vedant and Rueckert, Elmar},
	year = {2024},
	note = {arXiv:2401.17032 [cs]},
}

@article{garcia2008sensor,
  title={Sensor fusion for compliant robot motion control},
  author={Garc{\'\i}a, Javier G{\'a}mez and Robertsson, Anders and Ortega, Juan G{\'o}mez and Johansson, Rolf},
  journal={IEEE Transactions on Robotics},
  volume={24},
  number={2},
  pages={430--441},
  year={2008},
  publisher={IEEE}
}

@article{khalil2010dexterous,
  title={Dexterous robotic manipulation of deformable objects with multi-sensory feedback-a review},
  author={Khalil, Fouad F and Payeur, Pierre},
  journal={Robot Manipulators Trends and Development},
  number={March 2010},
  year={2010},
  publisher={InTech Croatia}
}

@article{xia2022review,
  title={A review on sensory perception for dexterous robotic manipulation},
  author={Xia, Ziwei and Deng, Zhen and Fang, Bin and Yang, Yiyong and Sun, Fuchun},
  journal={International Journal of Advanced Robotic Systems},
  year={2022},
  publisher={SAGE Publications Sage UK: London, England}
}

@article{alatise2020review,
  title={A review on challenges of autonomous mobile robot and sensor fusion methods},
  author={Alatise, Mary B and Hancke, Gerhard P},
  journal={IEEE Access},
  volume={8},
  pages={39830--39846},
  year={2020},
  publisher={IEEE}
}

@article{hu2016development,
  title={Development of sensory-motor fusion-based manipulation and grasping control for a robotic hand-eye system},
  author={Hu, Yingbai and Li, Zhijun and Li, Guanglin and Yuan, Peijiang and Yang, Chenguang and Song, Rong},
  journal={IEEE Transactions on Systems, Man, and Cybernetics: Systems},
  volume={47},
  number={7},
  pages={1169--1180},
  year={2016},
  publisher={IEEE}
}

@article{lee2020making,
  title={Making sense of vision and touch: Learning multimodal representations for contact-rich tasks},
  author={Lee, Michelle A and Zhu, Yuke and Zachares, Peter and Tan, Matthew and Srinivasan, Krishnan and Savarese, Silvio and Fei-Fei, Li and Garg, Animesh and Bohg, Jeannette},
  journal={IEEE Transactions on Robotics},
  volume={36},
  number={3},
  pages={582--596},
  year={2020},
  publisher={IEEE}
}

@article{elguea2023review,
  title={A review on reinforcement learning for contact-rich robotic manipulation tasks},
  author={Elguea-Aguinaco, {\'I}{\~n}igo and Serrano-Mu{\~n}oz, Antonio and Chrysostomou, Dimitrios and Inziarte-Hidalgo, Ibai and B{\o}gh, Simon and Arana-Arexolaleiba, Nestor},
  journal={Robotics and Computer-Integrated Manufacturing},
  volume={81},
  pages={102517},
  year={2023},
  publisher={Elsevier}
}

@inproceedings{li2021reinforcement,
  title={Reinforcement learning strategy based on multimodal representations for high-precision assembly tasks},
  author={Li, Ajian and Liu, Ruikai and Yang, Xiansheng and Lou, Yunjiang},
  booktitle={Intelligent Robotics and Applications: 14th International Conference, ICIRA 2021, Yantai, China, October 22--25, 2021, Proceedings, Part I 14},
  pages={56--66},
  year={2021},
  organization={Springer}
}

@article{suomalainen2022survey,
  title={A survey of robot manipulation in contact},
  author={Suomalainen, Markku and Karayiannidis, Yiannis and Kyrki, Ville},
  journal={Robotics and Autonomous Systems},
  volume={156},
  pages={104224},
  year={2022},
  publisher={Elsevier}
}

@article{liu2021deep,
  title={Deep reinforcement learning for the control of robotic manipulation: a focussed mini-review},
  author={Liu, Rongrong and Nageotte, Florent and Zanne, Philippe and de Mathelin, Michel and Dresp-Langley, Birgitta},
  journal={Robotics},
  volume={10},
  number={1},
  pages={22},
  year={2021},
  publisher={MDPI}
}

@article{levine2016end,
  title={End-to-end training of deep visuomotor policies},
  author={Levine, Sergey and Finn, Chelsea and Darrell, Trevor and Abbeel, Pieter},
  journal={Journal of Machine Learning Research},
  volume={17},
  number={39},
  pages={1--40},
  year={2016}
}

@article{whitney1987historical,
  title={Historical perspective and state of the art in robot force control},
  author={Whitney, Daniel E},
  journal={The International Journal of Robotics Research},
  volume={6},
  number={1},
  pages={3--14},
  year={1987},
  publisher={Sage Publications Sage CA: Thousand Oaks, CA}
}

@article{whitney1982quasi,
  title={Quasi-static assembly of compliantly supported rigid parts},
  author={Whitney, Daniel E and others},
  journal={Journal of Dynamic Systems, Measurement, and Control},
  volume={104},
  number={1},
  pages={65--77},
  year={1982},
  publisher={Citeseer}
}

@article{nuttin1997learning,
  title={Learning the peg-into-hole assembly operation with a connectionist reinforcement technique},
  author={Nuttin, Mamix and Van Brussel, H},
  journal={Computers in Industry},
  volume={33},
  number={1},
  pages={101--109},
  year={1997},
  publisher={Elsevier}
}

@inproceedings{feng2024play,
  title={Play to the Score: Stage-Guided Dynamic Multi-Sensory Fusion for Robotic Manipulation},
  author={Feng, Ruoxuan and Hu, Di and Ma, Wenke and Li, Xuelong},
  booktitle={8th Annual Conference on Robot Learning},
  year={2024},
}

@article{guzey2023dexterity,
  title={Dexterity from touch: Self-supervised pre-training of tactile representations with robotic play},
  author={Guzey, Irmak and Evans, Ben and Chintala, Soumith and Pinto, Lerrel},
  journal={arXiv preprint arXiv:2303.12076},
  year={2023}
}

@article{han2024learning,
  title={Learning generalizable vision-tactile robotic grasping strategy for deformable objects via transformer},
  author={Han, Yunhai and Yu, Kelin and Batra, Rahul and Boyd, Nathan and Mehta, Chaitanya and Zhao, Tuo and She, Yu and Hutchinson, Seth and Zhao, Ye},
  journal={IEEE/ASME Transactions on Mechatronics},
  year={2024},
  publisher={IEEE}
}

@misc{mnih_playing_2013,
	title = {Playing {Atari} with {Deep} {Reinforcement} {Learning}},
	url = {http://arxiv.org/abs/1312.5602},
	doi = {10.48550/arXiv.1312.5602},
	urldate = {2025-06-02},
	publisher = {arXiv},
	author = {Mnih, Volodymyr and Kavukcuoglu, Koray and Silver, David and Graves, Alex and Antonoglou, Ioannis and Wierstra, Daan and Riedmiller, Martin},
		year = {2013},
	note = {arXiv:1312.5602 [cs]},
}

@misc{zhang_towards_2015,
	title = {Towards {Vision}-{Based} {Deep} {Reinforcement} {Learning} for {Robotic} {Motion} {Control}},
	url = {http://arxiv.org/abs/1511.03791},
	doi = {10.48550/arXiv.1511.03791},
	urldate = {2025-06-02},
	publisher = {arXiv},
	author = {Zhang, Fangyi and Leitner, Jürgen and Milford, Michael and Upcroft, Ben and Corke, Peter},
		year = {2015},
	note = {arXiv:1511.03791 [cs]},
}

@misc{zhuang_robot_2023,
	title = {Robot {Parkour} {Learning}},
	url = {http://arxiv.org/abs/2309.05665},
	doi = {10.48550/arXiv.2309.05665},
	urldate = {2025-06-02},
	publisher = {arXiv},
	author = {Zhuang, Ziwen and Fu, Zipeng and Wang, Jianren and Atkeson, Christopher and Schwertfeger, Soeren and Finn, Chelsea and Zhao, Hang},
		year = {2023},
	note = {arXiv:2309.05665 [cs]},
}

@misc{garcin_studying_2025,
	title = {Studying the {Interplay} {Between} the {Actor} and {Critic} {Representations} in {Reinforcement} {Learning}},
	url = {http://arxiv.org/abs/2503.06343},
	doi = {10.48550/arXiv.2503.06343},
	urldate = {2025-06-25},
	publisher = {arXiv},
	author = {Garcin, Samuel and McInroe, Trevor and Castro, Pablo Samuel and Panangaden, Prakash and Lucas, Christopher G. and Abel, David and Albrecht, Stefano V.},
		year = {2025},
	note = {arXiv:2503.06343 [cs]},
}

@misc{luo2024serl,
    title={SERL: A Software Suite for Sample-Efficient Robotic Reinforcement Learning},
    author={Jianlan Luo and Zheyuan Hu and Charles Xu and You Liang Tan and Jacob Berg and Archit Sharma and Stefan Schaal and Chelsea Finn and Abhishek Gupta and Sergey Levine},
    year={2024},
    eprint={2401.16013},
    archivePrefix={arXiv},
    primaryClass={cs.RO}
}

@misc{chen_context_2023,
	title = {Context {Autoencoder} for {Self}-{Supervised} {Representation} {Learning}},
	url = {http://arxiv.org/abs/2202.03026},
	doi = {10.48550/arXiv.2202.03026},
	urldate = {2025-08-24},
	publisher = {arXiv},
	author = {Chen, Xiaokang and Ding, Mingyu and Wang, Xiaodi and Xin, Ying and Mo, Shentong and Wang, Yunhao and Han, Shumin and Luo, Ping and Zeng, Gang and Wang, Jingdong},
		year = {2023},
	note = {arXiv:2202.03026 [cs]},
}

@misc{ball_efficient_2023,
	title = {Efficient {Online} {Reinforcement} {Learning} with {Offline} {Data}},
	url = {http://arxiv.org/abs/2302.02948},
	doi = {10.48550/arXiv.2302.02948},
	urldate = {2025-08-24},
	publisher = {arXiv},
	author = {Ball, Philip J. and Smith, Laura and Kostrikov, Ilya and Levine, Sergey},
	year = {2023},
	note = {arXiv:2302.02948 [cs]},
}

\end{document}